\documentstyle[proc]{article}

\author{Marc Dymetman\thanks{Thanks to Pierre Isabelle and François
Perrault for their comments, and to CITI (Montreal) 
for its support during the preparation of this paper.} 
\\[1ex]
Rank Xerox Research Centre \\ 6, chemin de Maupertuis \\ Meylan, 38240
France \\{\tt dymetman@xerox.fr}}
\title{A Simple Transformation for Offline-Parsable Grammars and its
Termination Properties}

\date{}

\def\key#1{{\bf #1}}

\def\el{[\hspace*{1mm}]}
\def\check{$\times$}
\def\myindent{\hspace*{1.5mm}}
\def\ione{}
\def\itwo{\myindent\myindent}
\def\ithree{\myindent\myindent\myindent}
\def\ifour{\myindent\myindent\myindent\myindent}
\def\ifive{\myindent\myindent\myindent\myindent\myindent}
\def\singlestrut{\vrule width 0pt height 8.5pt depth 3.5pt}

\long\def\framedminipage#1\endframedminipage{{\makebox[1.0\textwidth][c]%
{\begin{minipage}{1.0\textwidth}\small\sf\parindent1em\vspace{.25cm}%
\begingroup#1\endgroup\vspace{.25cm}\end{minipage}}}}

\begin{document}

\maketitle
\bibliographystyle{plain}

\paragraph{Abstract} We present, in easily reproducible terms, 
a simple transformation for offline-parsable grammars which results in
a provably terminating parsing program directly top-down interpretable
in Prolog.  The transformation consists in two steps: (1) removal of
empty-productions, followed by: (2) left-recursion elimination. It is
related both to left-corner parsing (where the grammar is compiled,
rather than interpreted through a parsing program, and with the
advantage of guaranteed termination in the presence of empty
productions) and to the Generalized Greibach Normal Form for DCGs
(with the advantage of implementation simplicity).

\section{Motivation}

Definite clause grammars (DCGs) are one of the simplest and most
widely used unification grammar formalisms. They represent a direct
augmentation of context-free grammars through the use of (term)
unification (a fact that tends to be masked by their usual
presentation based on the programming language Prolog). It is
obviously important to ask wether certain usual methods and algorithms
pertaining to CFGs can be adapted to DCGs, and this general question
informs much of the work concerning DCGs, as well as more complex
unification grammar formalisms (to cite only a few areas: Earley
parsing, LR parsing, left-corner parsing, Greibach Normal Form). 

One essential complication when trying to generalize CFG methods to
the DCG domain lies in the fact that, whereas the parsing problem for
CFGs is decidable, the corresponding problem for DCGs is in general
undecidable. This can be shown easily as a consequence of the
noteworthy fact that any definite clause {\em program} can be viewed
as a definite clause {\em grammar} ``on the empty string'', that is,
as a DCG where no terminals other than $\el$ are allowed on the
right-hand sides of rules.  The Turing-completeness of definite clause
programs therefore implies the undecidability of the parsing problem
for this subclass of DCGs, and {\em a fortiori} for DCGs in
general.%
\footnote{DCGs on the empty string might be dismissed as
extreme, but they are in fact at the core of the offline-parsability
concept. See note \ref{ggnf}.}
 In order to guarantee good computational properties for DCGs, it is
then necessary to impose certain restrictions on their form such as
{\em offline-parsability} (OP), a nomenclature introduced by Pereira
and Warren \cite{PeWa83}, who define an OP DCG as a grammar whose
context-free skeleton CFG is not infinitely ambiguous, and show that
OP DCGs lead to a decidable parsing problem.\footnote{The concept of
offline-parsability (under a different name) goes back to
\cite{KaBr82}, where it is shown to be linguistically relevant.}

Our aim in this paper is to propose a {\em simple} transformation for
an arbitrary OP DCG putting it into a form which leads to the
completeness of the direct top-down interpretation by the standard
Prolog interpreter: parsing is guaranteed to enumerate all solutions
to the parsing problem and terminate. The existence of such a
transformation is known: in \cite{DymColing92,Dy91}, we
have recently introduced a ``Generalized Greibach Normal Form'' (GGNF)
for DCGs, which leads to termination of top-down interpretation in the
OP case. However, the available presentation of the GGNF
transformation is rather complex (it involves an algebraic study of
the fixpoints of certain equational systems representing
grammars.). Our aim here is to present a related, but much simpler,
transformation, which from a theoretical viewpoint performs somewhat
less than the GGNF transformation (it involves some encoding of the
initial DCG, which the GGNF does not, and it only handles
offline-parsable grammars, while the GGNF is defined for arbitrary
DCGs),%
\footnote{\label{ggnf}%
The GGNF factorizes an arbitrary DCG into two
components: a ``unit sub-DCG on the empty string'', and another part
consisting of rules whose right-hand side starts with a terminal. The
decidability of the DCG depends exclusively on certain simple textual
properties of the unit sub-DCG. This sub-DCG can be eliminated from
the GGNF if and only if the DCG is offline-parsable.}
but in practice is
extremely easy to implement and displays a comparable behavior when
parsing with an OP grammar.

The transformation consists of two steps: (1) empty-production
elimination and (2) left-recursion elimination.

The empty-production elimination algorithm is inspired by the
usual procedure for context-free grammars. But there are some notable
differences, due to the fact that removal of empty-productions is in
general impossible for non-OP DCGs. The empty-production elimination
algorithm is guaranteed to terminate only in the OP case.%
\footnote{The fact that the standard CFG empty-production elimination 
transformation is always possible is related to the fact that this
transformation does not preserve degrees of ambiguity. For instance
the infinitely ambiguous grammar $S \rightarrow [b]\: A,\: A
\rightarrow A,\: A \rightarrow \el$ is simplified into the 
grammar $S \rightarrow [b]$.  This type of simplification is generally
impossible in a DCG. Consider for instance the ``grammar'' $s(X)
\rightarrow [number]\: a(X),\: a(succ(X)) \rightarrow a(X),\: a(0)
\rightarrow \el$.}
It produces a DCG declaratively equivalent to the original grammar.

The left-recursion elimination algorithm is adapted from a
transformation proposed in \cite{DyIsPe90} in the context of a certain
formalism (``Lexical Grammars'') which we presented as a
possible basis for building reversible grammars.%
\footnote{The method goes back to a transformation used to compile out
certain local cases of left-recursion from DCGs in the context of the
Machine Translation prototype CRITTER \cite{DyIs88}.}
The key observation (in slightly different terms) was that, in a DCG,
if a nonterminal $g$ is defined literally by the two rules (the first
of which is left-recursive):
\begin{eqnarray*}
&&      g(X) \rightarrow g(Y),\: d(Y,X).                \\
&&      g(X) \rightarrow t(X).                          
\end{eqnarray*}
then the replacement of these two rules by the three rules (where
$d\_tc$ is a new nonterminal symbol, which represents a kind of
``transitive closure'' of $d$):
\begin{eqnarray*}
&&      g(X) \rightarrow t(Y),\: d\_tc(Y,X).            \\
&&      d\_tc(X,X) \rightarrow \el.                     \\
&&      d\_tc(X,Z) \rightarrow d(X,Y),\: d\_tc(Y,Z).    
\end{eqnarray*}
preserves the declarative semantics of the grammar.\footnote{A proof
of this fact, based on a comparison of proof-trees for the original
and the transformed grammar, is given in \cite{Dy91}.}

We remarked in \cite{DyIsPe90} that this
transformation ``is closely related to left-corner parsing'', but did
not give details. In a recent paper \cite{Johnson:Left-corner}, Mark
Johnson introduces ``a left-corner program transformation for natural
language parsing'', which has some similarity to the above
transformation, but which is applied to definite clause programs,
rather than to DCGs. He proves that this transformation respects
declarative equivalence, and also shows, using a model-theoretic
approach, the close connection of his transformation with left-corner
parsing \cite{RosencrantzLeftCorner,MaTaHiMiYa83,PeSh87}.%
\footnote{
His paper does not state termination conditions for the transformed
program. Such termination conditions would probably involve some
generalized notion of offline-parsability
\cite{Jo88,Haas89,Sh92}. By contrast, we prove
termination only for DCGs which are OP in the original sense of
Pereira and Warren, but this case seems to us to represent much of the
core issue, and to lead to some direct extensions. For instance, the
DCG transformation proposed here can be directly applied to ``guided''
programs in the sense of \cite{DyIsPe90}.}

It must be noted that the left-recursion elimination procedure can be
applied to any DCG, whether OP or not. Even in the case where the
grammar is OP, however, it will {\em not} lead to a terminating
parsing algorithm {\em unless} empty productions have been prealably
eliminated from the grammar, a problem which is shared by the usual
left-corner parser-interpreter.

{\em Due to the space available, we do not give here correctness
proofs for the algorithms presented, but expect to publish them in a
fuller version of this paper.  These algorithms have actually been
implemented in a slightly extended version, where they are also used
to decide whether the grammar proposed for transformation is in fact
offline-parsable or not.}

\mbox{\rule{0mm}{0mm}} 

\vspace*{-8mm}

\section{Empty-production elimination}

It can be proven that, if DCG0 is an OP DCG, the following transformation,
which involves repeated partial evaluation of rules that rewrite into
the empty string, terminates after a finite number of steps and
produces a grammar DCG without empty-productions which is equivalent
to the initial grammar on non-empty strings:%
\footnote{When DCG0 is not
OP, the transformation may produce an infinite number of rules, but a
simple extension of the algorithm can detect this situation: the
transformation stops and the grammar is declared not to be OP.}

\mbox{\rule{0mm}{0mm}} 

\vspace*{-1.3ex}

{\small
\offinterlineskip
\tabskip=0mm
\halign{\singlestrut#\hfill\cr
\ione \key{input}: an offline-parsable DCG1. \cr
\ione \key{output}: a DCG without empty rules equivalent to DCG1 \cr
\ione on non-empty strings. \cr
\ione \key{algorithm}:  \cr
\itwo initialize  LIST1 to a list of the rules of DCG1, set LIST2 \cr
\itwo to the empty list. \cr
\itwo \key{while} there exists an empty rule ER: \enskip $A(T1,...,Tk)
\rightarrow \el$ \cr
\itwo in LIST1 \key{do:} \cr
  \ithree move ER to LIST2. \cr
  \ithree \key{for} each rule R: \enskip $B(...) \rightarrow \alpha$ \enskip in LIST1 such that $\alpha$ \cr
  \ithree contains an instance of $A(...)$ (including \cr
  \ithree new such rules created inside this loop) \key{do:}  \cr
    \ifour \key{for} each such instance $A(S1,...,Sk)$ unifiable with \cr 
    \ifour $A(T1,...,Tk)$ \key{do:} \cr
      \ifive append to LIST1 a rule R': \enskip $B(...) \rightarrow \alpha'$ \enskip obtained \cr
      \ifive from R by removing $A(S1,...,Sk)$  \cr
      \ifive from $\alpha$ (or by replacing it with $\el$ if this was \cr
      \ifive the only nonterminal in $\alpha$),  \cr
      \ifive and by unifying the $Ti$'s with the $Si$'s. \cr
\itwo set DCG to LIST1. \cr
}}

\rule{0mm}{1pt} 
\vspace*{-.2cm}

For instance the grammar consisting in the nine rules appearing above
the separation in fig.~\ref{Removal of empty rules.} is transformed into
the grammar (see figure):

\rule{0mm}{1pt} 

\vspace*{-1ex}

{\small
\offinterlineskip
\halign{\rule{2cm}{0mm}\rule{0mm}{4mm}$#$\hfil\cr
s(s(NP,VP)) \rightarrow np(NP), vp(VP). \cr
np(np(N,C)) \rightarrow n(N), comp(C). \cr
n(n(people)) \rightarrow [people]. \cr
vp(vp(v(sleep),C)) \rightarrow [sleep], comp(C). \cr
comp(c(C,A)) \rightarrow comp(C), adv(A). \cr
adv(adv(here)) \rightarrow [here]. \cr
adv(adv(today)) \rightarrow [today]. \cr
np(np(n(you)),C) \rightarrow comp(C). \cr
np(np(N,nil)) \rightarrow n(N). \cr
comp(c(nil,A)) \rightarrow adv(A). \cr
vp(vp(v(sleep),nil)) \rightarrow [sleep]. \cr
s(s(np(np(n(you)),nil),VP)) \rightarrow vp(VP). \cr
}}

\section{Left-recursion elimination}

The transformation can be logically divided into two steps: (1) an
encoding of DCG into a ``generic'' form DCG', and (2) a simple
replacement of a certain group of left-recursive rules in DCG' by a
certain equivalent non left-recursive group of rules, yielding a
top-down interpretable DCG''. An example of the transformation DCG $\longrightarrow$
DCG' $\longrightarrow$ DCG'' is given in fig.~\ref{Encoding and left-recursion
elimination}. 

The encoding is performed by the following algorithm:

\mbox{\rule{1mm}{0mm}} 
\vspace*{-.4 cm}

{\small
\offinterlineskip
\tabskip=0mm
\halign{\singlestrut#\hfill\cr
\ione  \cr
\ione \key{input}: an offline-parsable DCG without empty rules. \cr
\ione \key{output}: an equivalent ``encoding'' DCG'. \cr
\ione \key{algorithm}:  \cr
\itwo initialize  LIST to a list of the rules of DCG. \cr
\itwo initialize  DCG' to the list of rules (literally):  \cr
\itwo \quad $g(X) \rightarrow g(Y),\; d(Y,X).$  \cr
\itwo \quad $g(X) \rightarrow t(X).$  \cr
\itwo \key{while} there exists a rule R of the form \cr 
\itwo $A(T1,...,Tk) \rightarrow B(S1,...,Sl)\; \alpha$  in LIST \key{do:} \cr
  \ithree remove R from LIST. \cr
  \ithree add to DCG' a rule R': \cr
  \ithree $d(B(S1,...,Sl),A(T1,...,Tk)) \rightarrow \alpha'$, \cr
  \ithree where $\alpha'$ is obtained by replacing any $C(V1,...,Vm)$ \cr
  \ithree in $\alpha$ by $g(C(V1,...,Vm))$, \cr
  \ithree or is set to $\el$ in the case where $\alpha$ is empty. \cr
\itwo \key{while} there exists a rule R of the form \cr 
\itwo $A(T1,...,Tk) \rightarrow [terminal]\; \alpha$ \enskip in LIST \key{do:} \cr
  \ithree remove R from LIST. \cr
  \ithree add to DCG' a rule R': \cr
  \ithree  $t(A(T1,...,Tk)) \rightarrow [terminal]\; \alpha'$, \cr
  \ithree where $\alpha'$ is obtained by replacing any $C(V1,...,Vm)$ \cr
  \ithree in $\alpha$ by $g(C(V1,...,Vm))$, \cr
  \ithree or is set to $\el$ in the case where $\alpha$ is empty. \cr
}}

\mbox{\rule{1mm}{0mm}} 
\vspace*{.1 cm}

The procedure is very simple. It involves the creation of a generic
nonterminal $g(X)$, of arity one, which performs a task equivalent to
the original nonterminals $s(X1,\ldots,Xn), vp(X1,\ldots,Xm),
...$. The goal $g(s(X1,\ldots,Xn))$, for instance, plays the same role
for parsing a sentence as did the goal $s(X1,\ldots,Xn)$ in the
original grammar.

Two further generic nonterminals are introduced: $t(X)$ accounts for
rules whose right-hand side begins with a terminal, while
$d(Y,X)$ accounts for rules whose right-hand side begins with a
nonterminal. The rationale behind the encoding is best understood from
the following examples, where $\Longrightarrow$ represents rule
rewriting:\\[2mm]
$vp(vp(v(sleep),C)) \rightarrow [sleep],\: comp(C)                                              \\
\Longrightarrow  g(vp(vp(v(sleep),C))) \rightarrow [sleep],\: g(comp(C))        \\
\Longrightarrow  g(X) \rightarrow [sleep],\\ \mbox{\quad} \underbrace{(\;\{X=vp(vp(v(sleep),C))\},\; g(comp(C))\;)}_{t(X)} $

\newpage
\noindent $s(s(NP,VP)) \rightarrow np(NP),\: vp(VP) \\
\Longrightarrow  g(s(s(NP,VP))) \rightarrow g(np(NP)),\: g(vp(VP)) \\
\Longrightarrow  g(X) \rightarrow g(Y),\\ \mbox{\quad} \underbrace{(\;\{X=s(s(NP,VP)),\: Y=np(NP)\},\; g(vp(VP))\;)}_{d(Y,X)}$


The second example illustrates the role played by $d(Y,X)$ in the
encoding. This nonterminal has the following interpretation: $X$ is
an``immediate'' extension of $Y$ using the given rule. In other words,
$Y$ corresponds to an ``immediate left-corner'' of $X$.

The left-recursion elimination is now performed by the following
``algorithm'':\footnote{In practice, this and the preceding algorithm,
which are dissociated here for exposition reasons, are 
lumped together.}

\mbox{\rule{1mm}{0mm}} 
\vspace*{-.4 cm}

{\small
\offinterlineskip
\tabskip=0mm
\halign{\singlestrut#\hfill\cr
\ione  \cr
\ione \key{input}: a DCG' encoded as above.\cr
\ione \key{output}: an equivalent non left-recursive DCG''. \cr
\ione \key{algorithm}:  \cr
\itwo initialize  DCG'' to DCG'.  \cr
\itwo in DCG'', replace literally the rules:  \cr
\itwo \quad $g(X) \rightarrow g(Y),\; d(Y,X).$  \cr
\itwo \quad $g(X) \rightarrow t(X).$  \cr
\itwo by the rules:  \cr
\itwo \quad $g(X) \rightarrow t(Y),\; d\_tc(Y,X).$  \cr
\itwo \quad $d\_tc(X,X) \rightarrow \el.$ \cr
\itwo \quad $d\_tc(X,Z) \rightarrow d(X,Y),\; d\_tc(Y,Z).$  \cr
}}

\mbox{\rule{1mm}{0mm}} 
\vspace*{.1 cm}

In this transformation, the new nonterminal $d\_tc$ plays the role of
a kind of transitive closure of $d$. It can be seen that, relative to
DCG'', for any string $w$ and for any ground term $z$, the
fact that 
\enskip $g(z)$ \enskip rewrites into $w$ ---or, equivalently, that
there exists a ground term $x$ such that \enskip $t(x)\:d\_tc(x,z)$
\enskip rewrites into $w$---is equivalent to the existence of a
sequence of ground terms $x=x_1,\; \ldots,\; x_k=z$ and a sequence of
strings $w_1,\; \ldots,\; w_k$ such that $t(x_1)$ rewrites into $w_1$,
\enskip $d(x_1,x_2)$ rewrites into $w_2$,\enskip ..., \enskip
$d(x_{k-1},x_k)$ rewrites into $w_k$, and such that $w$ is the string
concatenation $w = w_1 \cdots w_k$. From our previous remark on the
meaning of $d(Y,X)$, this can be interpreted as saying that
``consituent $x$ is a left-corner of constituent $z$'', relatively to
string $w$.

The grammar DCG'' can now be compiled in the standard way---via the
adjunction of two ``differential list'' arguments---into a Prolog
program which can be executed directly. If we started from an
offline-parsable grammar DCG0, this program will enumerate all
solutions to the parsing problem and terminate after a finite number
of steps.%
\footnote{There exist of course DCGs which do not contain empty
productions and which are {\em not} OP. They are characterized by the
existence of cycles of ``chain-rules'' of the form: $a_1(...)
\rightarrow a_2(...).$ , ...,  $a_{m-1}(...) \rightarrow a_m(...).$ , with $a_m
= a_1$. But, if we start with an OP DCG0, the empty-production
elimination algorithm cannot produce such a situation.}

\begin{figure*}[t]
\framedminipage
{\small
\offinterlineskip
\tabskip=8mm
\halign{\rule{0mm}{4mm}$#$\hfil&#&$#$\hfill\cr
\mbox{\rm LIST1} &\rm delete & \rm LIST2 \cr
&& \cr
s(s(NP,VP)) \rightarrow np(NP), vp(VP). & & \cr
np(np(N,C)) \rightarrow n(N), comp(C). & & \cr
n(n(people)) \rightarrow [people]. & & \cr
n(n(you)) \rightarrow \el. & \check & \cr
vp(vp(v(sleep),C)) \rightarrow [sleep], comp(C). & & \cr
comp(c(C,A)) \rightarrow comp(C), adv(A). & & \cr
comp(nil) \rightarrow \el. & \check & \cr
adv(adv(here)) \rightarrow [here]. & & \cr
adv(adv(today)) \rightarrow [today]. & & \cr
\rule{0mm}{6mm} \hrulefill & &  \cr
 & & n(n(you)) \rightarrow \el. \cr
np(np(n(you)),C) \rightarrow comp(C). & & \cr
 & & comp(nil) \rightarrow \el. \cr
np(np(N,nil)) \rightarrow n(N). & & \cr
comp(c(nil,A)) \rightarrow adv(A). & & \cr
vp(vp(v(sleep),nil)) \rightarrow [sleep]. & & \cr
np(np(n(you)),nil) \rightarrow \el. & \check & \cr
 &  & np(np(n(you)),nil) \rightarrow \el. \cr
s(s(np(np(n(you)),nil),VP)) \rightarrow vp(VP). & & \cr
}}
\endframedminipage
\caption{Empty-production elimination.} 
\label{Removal of empty rules.}
\end{figure*}

\begin{figure*}[p]
\framedminipage
{\small
\offinterlineskip
\tabskip=10mm
\halign{\rule{0mm}{4mm}$#$\hfil & $#$\hfil \cr
\mbox{\rm DCG}                                    &    \mbox{\rm DCG'} \cr                             \cr
                                                  &    g(X) \rightarrow g(Y), d(Y,X).                                \cr
                                                  &    g(X) \rightarrow t(X).                                        \cr
s(s(NP,VP)) \rightarrow np(NP), vp(VP).           &    d(np(NP),s(s(NP,VP))) \rightarrow g(vp(VP)).            \cr
np(np(N,C)) \rightarrow n(N), comp(C).            &    d(n(N),np(np(N,C))) \rightarrow g(comp(C)).             \cr
n(n(people)) \rightarrow [people].                &    t(n(n(people))) \rightarrow [people].                   \cr
vp(vp(v(sleep),C)) \rightarrow [sleep], comp(C).  &    t(vp(vp(v(sleep),C))) \rightarrow [sleep], g(comp(C)).  \cr
comp(c(C,A)) \rightarrow comp(C), adv(A).         &    d(comp(C),comp(c(C,A))) \rightarrow g(adv(A)).          \cr
adv(adv(here)) \rightarrow [here].                &    t(adv(adv(here))) \rightarrow [here].                   \cr
adv(adv(today)) \rightarrow [today].              &    t(adv(adv(today))) \rightarrow [today].                 \cr
np(np(n(you)),C) \rightarrow comp(C).             &    d(comp(C),np(np(n(you)),C)) \rightarrow \el.             \cr
np(np(N,nil)) \rightarrow n(N).                   &    d(n(N),np(np(N,nil))) \rightarrow \el.                   \cr
comp(c(nil,A)) \rightarrow adv(A).                &    d(adv(A),comp(c(nil,A))) \rightarrow \el.                \cr
vp(vp(v(sleep),nil)) \rightarrow [sleep].         &    t(vp(vp(v(sleep),nil))) \rightarrow [sleep].            \cr
s(s(np(np(n(you)),nil),VP)) \rightarrow vp(VP).   &    d(vp(VP),s(s(np(np(n(you)),nil),VP))) \rightarrow \el.   \cr
                                                  &                                                  \cr
                                                  &                                                  \cr
                                                  &    \mbox{\rm DCG''}                              \cr
                                                  &                                                  \cr
                                                  &    g(X) \rightarrow t(Y), d\_tc(Y,X).                                \cr
                                                  &    d\_tc(X,X) \rightarrow \el.                                        \cr
                                                  &    d\_tc(X,Z) \rightarrow d(X,Y), d\_tc(Y,Z).                          \cr
                                                  &    d(np(NP),s(s(NP,VP))) \rightarrow g(vp(VP)).                    \cr
                                                  &    d(n(N),np(np(N,C))) \rightarrow g(comp(C)).                     \cr
                                                  &    t(n(n(people))) \rightarrow [people].                           \cr
                                                  &    t(vp(vp(v(sleep),C))) \rightarrow [sleep], g(comp(C)).          \cr
                                                  &    d(comp(C),comp(c(C,A))) \rightarrow g(adv(A)).                  \cr
                                                  &    t(adv(adv(here))) \rightarrow [here].                           \cr
                                                  &    t(adv(adv(today))) \rightarrow [today].                         \cr
                                                  &    d(comp(C),np(np(n(you)),C)) \rightarrow \el.                     \cr
                                                  &    d(n(N),np(np(N,nil))) \rightarrow \el.                           \cr
                                                  &    d(adv(A),comp(c(nil,A))) \rightarrow \el.                        \cr
                                                  &    t(vp(vp(v(sleep),nil))) \rightarrow [sleep].                    \cr
                                                  &    d(vp(VP),s(s(np(np(n(you)),nil),VP))) \rightarrow \el.   \cr
}}
\endframedminipage
\caption{Encoding (DCG') of a grammar (DCG) and left-recursion elimination (DCG'').} 
\label{Encoding and left-recursion elimination}
\end{figure*}


\begin{thebibliography}{10}

\bibitem{DymColing92}
Marc Dymetman.
\newblock {A Generalized Greibach Normal Form for Definite Clause Grammars}.
\newblock In {\em Proceedings of the 15th International Conference on
  Computational Linguistics}, volume~1, pages 366--372, Nantes, France, July
  1992.

\bibitem{Dy91}
Marc Dymetman.
\newblock {Transformations de grammaires logiques et r\'eversibilit\'e en
  Traduction Automatique}.
\newblock Th\`ese d'Etat, 1992.
\newblock Universit\'e Joseph Fourier (Grenoble 1), Grenoble, France.

\bibitem{DyIs88}
Marc Dymetman and Pierre Isabelle.
\newblock Reversible logic grammars for machine translation.
\newblock In {\em Proceedings of the Second International Conference on
  Theoretical and Methodological Issues in Machine Translation of Natural
  Languages}, Pittsburgh, PA, June 1988. Carnegie Mellon University.

\bibitem{DyIsPe90}
Marc Dymetman, Pierre Isabelle, and Fran\c{c}ois Perrault.
\newblock A symmetrical approach to parsing and generation.
\newblock In {\em Proceedings of the 13th International Conference on
  Computational Linguistics}, volume~3, pages 90--96, Helsinki, August 1990.

\bibitem{Haas89}
Andrew Haas.
\newblock A generalization of the offline-parsable grammars.
\newblock In {\em Proceedings of the 27th Annual Meeting of the Association for
  Computational Linguistics}, pages 237--42, Vancouver, BC, Canada, June 1989.

\bibitem{Jo88}
Mark Johnson.
\newblock {\em Attribute-Value Logic and the Theory of Grammar}.
\newblock CSLI lecture note No. 16. Center for the Study of Language and
  Information, Stanford, CA, 1988.

\bibitem{Johnson:Left-corner}
Mark Johnson.
\newblock A left-corner program transformation for natural language parsing,
  (forthcoming).

\bibitem{KaBr82}
R.~Kaplan and J.~Bresnan.
\newblock Lexical functional grammar: a formal system for grammatical
  representation.
\newblock In Bresnan, editor, {\em The Mental Representation of Grammatical
  Relations}, pages 173--281. MIT Press, Cambridge, MA, 1982.

\bibitem{MaTaHiMiYa83}
Y.~Matsumoto, H.~Tanaka, H.~Hirikawa, H.~Miyoshi, and H.~Yasukawa.
\newblock {BUP}: a bottom-up parser embedded in {Prolog}.
\newblock {\em New Generation Computing}, 1(2):145--158, 1983.

\bibitem{PeSh87}
Fernando C.~N. Pereira and Stuart~M. Shieber.
\newblock {\em Prolog and Natural Language Analysis}.
\newblock CSLI lecture note No. 10. Center for the Study of Language and
  Information, Stanford, CA, 1987.

\bibitem{PeWa83}
Fernando C.~N. Pereira and David H.~D. Warren.
\newblock Parsing as deduction.
\newblock In {\em Proceedings of the 21th Annual Meeting of the Association for
  Computational Linguistics}, pages 137--144, MIT, Cambridge, MA, June 1983.

\bibitem{RosencrantzLeftCorner}
D.~J. Rosencrantz and P.~M. Lewis.
\newblock Deterministic left-corner parsing.
\newblock In {\em Eleventh Annual Symposium on Switching and Automata Theory},
  pages 139--153. IEEE, 1970.
\newblock Extended Abstract.

\bibitem{Sh92}
Stuart~M. Shieber.
\newblock {\em Constraint-Based Grammar Formalisms}.
\newblock MIT Press, Cambridge, MA, 1992.

\end{thebibliography}
\end{document}